\documentclass[10pt,twocolumn,letterpaper]{article}

\usepackage{cvpr}
\usepackage{times}
\usepackage{epsfig}
\usepackage{graphicx}
\usepackage{amsmath}
\usepackage{amssymb}

\usepackage{subfigure}
\usepackage{caption}  
\usepackage{epstopdf}
\usepackage{url}
\usepackage[ruled, linesnumbered, noend]{algorithm2e}
\usepackage{booktabs}
\usepackage{multirow}
\usepackage{multicol}

\usepackage[breaklinks=true,bookmarks=false]{hyperref}

\cvprfinalcopy 

\def\cvprPaperID{****} 

\begin{document}

\title{Deep Active Learning for Video-based Person Re-identification}

\author{Menglin Wang$^1$,
Baisheng Lai$^2$,
Zhongming Jin$^2$,
Xiaojin Gong$^1$,
Jianqiang Huang$^2$,
Xiansheng Hua$^2$
\\
$^1$ Zhejiang University;  \  $^2$ Alibaba Group\\
{\tt\small \{menglinwang, gongxj\}@zju.edu.cn}\\
{\tt\small \{baisheng.lbs, zhongming.jinzm, jianqiang.hjq, xiansheng.hxs\}@alibaba-inc.com}\\
}

\maketitle

\begin{abstract}
It is prohibitively expensive to annotate a large-scale video-based person re-identification (re-ID) dataset, which makes fully supervised methods inapplicable to real-world deployment. How to maximally reduce the annotation cost while retaining the re-ID performance becomes an interesting problem. In this paper, we address this problem by integrating an active learning scheme into a deep learning framework. Noticing that the truly matched tracklet-pairs, also denoted as true positives (TP), are the most informative samples for our re-ID model, we propose a sampling criterion to choose the most TP-likely tracklet-pairs for annotation. A view-aware sampling strategy considering view-specific biases is designed to facilitate candidate selection, followed by an adaptive resampling step to leave out the selected candidates that are unnecessary to annotate. Our method learns the re-ID model and updates the annotation set iteratively. The re-ID model is supervised by the tracklets' pesudo labels that are initialized by treating each tracklet as a distinct class. With the gained annotations of the actively selected candidates, the tracklets' pesudo labels are updated by label merging and further used to re-train our re-ID model. While being simple, the proposed method demonstrates its effectiveness on three video-based person re-ID datasets. Experimental results show that less than 3\% pairwise annotations are needed for our method to reach comparable performance with the fully-supervised setting.  
\end{abstract}

\section{Introduction}
\label{sec:introduction}

Person re-identification (re-ID) in video surveillance has important significance for public security. Therefore, extensive studies have been conducted to address the re-ID problem and most of them focused on single image frames~\cite{Ma2015image, Zheng2016Survey, Yu2017image, Liao2015image, Xiao2016image}. In recent years, video-based re-ID~\cite{chen2018video, li2017learning, li2017video, McLaughlin2016video, You2016video, Wang2014video} has been attracting more and more research attention. It utilizes both spatial and temporal information so that can better overcome the challenges resulted from occlusion, lighting variation, and pose and camera-view change. 

Most existing works perform video-based person re-ID under full supervision. While state-of-the-art re-ID performances are reported in large-scale labeled datasets~\cite{zheng2016mars, ristani2016performance}, the fully supervised methods~\cite{chen2018video, li2017learning, li2017video, McLaughlin2016video, You2016video} are weak when scaled to real-world deployment. The reasons are in two aspects. On one hand, the amount of video data collected by a wide-area camera network is large, it is prohibitively expensive to make full annotations. On the other hand, the abundant unlabeled data are informative but fully supervised methods rarely discover their inherent information. Researchers therefore resort to unsupervised~\cite{Yu2017image, ye2017dynamic, liu2017stepwise} or semi-supervised~\cite{wu2018exploit, liu2018semi} techniques. Unfortunately, there is still a significant performance gap between these methods and fully supervised counterparts so far.

In order to reduce the annotation cost while keeping the re-ID performance, this paper proposes an approach that integrates an active learning (AL)~\cite{Settles2009active} scheme into a deep learning framework. Active learning aims to use as few labeled data as possible to achieve high performance. The sampling strategy, that is how to pick the most informative instances for annotation, plays a key role. Different query strategies have been recently developed in various AL-based vision tasks such as classification~\cite{Li2014active}, recognition~\cite{Hasan2015active, Lin2018active}, and object detection~\cite{Roy2018active, Wan2018active}. However, they cannot be straightforwardly applied to the person re-ID problem because they do not exploit the inter-relations between samples but consider individual instances only. In recent years, there have also been several attempts at utilizing active learning for person re-ID~\cite{Das2015active, wang2016inloop, liu2013human, roy2018exploiting}. Some~\cite{Das2015active, wang2016inloop, liu2013human} focused on post-ranking and exploited the annotations to refine the initial ranking results. ~\cite{roy2018exploiting} explicitly considered AL in person re-ID as an optimal subset selection task and implemented it by solving a triangle free subgraph maximization problem on the k-partite graph. Few of these methods exploit the inter-relations between samples to facilitate sample selection and model learning.

In video-based person re-ID, the annotation task is either directly assigning an ID label to each tracklet or telling whether two tracklets are matched or not. In this work, we take the second annotation manner. By checking a common video-based person re-ID dataset, we notice that only a rather small portion of tracklet-pairs are true matches, also referred to as true positives (TP), and most pairs are negative. It indicates that the truly matched tracklet-pairs are the most informative candidates for learning. Motivated by this observation, active learning is exploited to find and annotate the most TP-likely tracklet-pairs that the re-ID model is certain of. This sampling criterion is distinct from typical active learning methods~\cite{Li2014active, liu2013human, Hasan2015active, wang2016inloop, Das2015active} in which the most uncertain samples are queried.

\begin{figure}[t]
\begin{minipage}[t]{0.15\textwidth}
\centering
\includegraphics[width=0.9\textwidth]{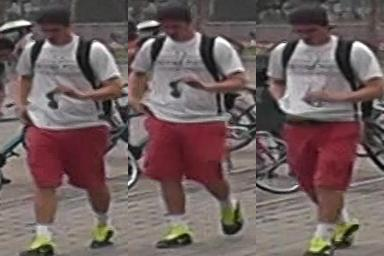}
\caption*{(a)}
\end{minipage} 
\begin{minipage}[t]{0.15\textwidth}
\centering
\includegraphics[width=0.9\textwidth]{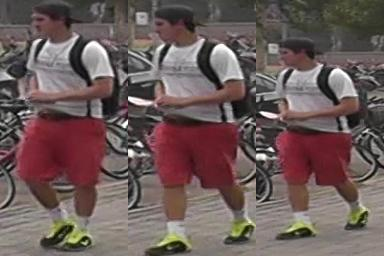}
\caption*{(b)}
\end{minipage} 
\begin{minipage}[t]{0.15\textwidth}
\centering
\includegraphics[width=0.9\textwidth]{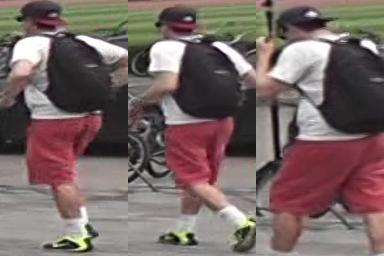}
\caption*{(c)}
\end{minipage} \\\hspace{0.5\linewidth}
\begin{minipage}[t]{0.15\textwidth}
\centering
\includegraphics[width=0.9\textwidth]{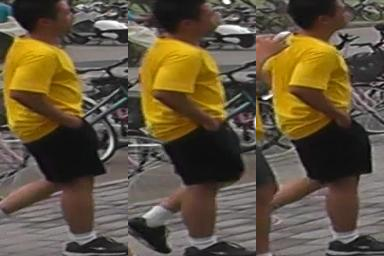}
\caption*{(d)}
\end{minipage} 
\begin{minipage}[t]{0.15\textwidth}
\centering
\includegraphics[width=0.9\textwidth]{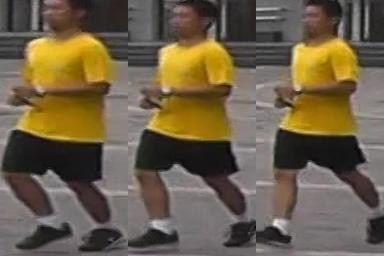}
\caption*{(e)}
\end{minipage} 
\begin{minipage}[t]{0.15\textwidth}
\centering
\includegraphics[width=0.9\textwidth]{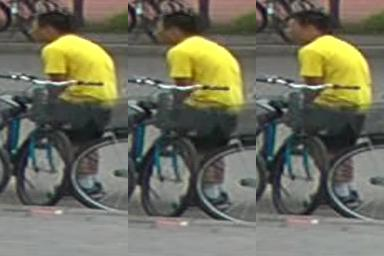}
\caption*{(f)}
\end{minipage}
\caption{Example of identities in the same and different camera views. Each row shows three tracklets of a person. In the first row, tracklets (a) and (b) are from the same camera, (c) is from a different camera. In the second row, tracklets (d) and (e) are from the same camera, (f) is from a different camera.}	
\label{camera-view}
\end{figure}

In addition, we also observe the following view-specific biases: 1) Truly matched tracklets from the same camera view are more similar to each other than those from different views, as shown in Fig.~\ref{camera-view}; 2) False positives are more likely to be selected from the same views. This observation inspires us to design a view-aware sampling strategy that takes the view information into account. An adaptive resampling step is further adopted to filter out the selected negative pairs that are unnecessary to annotate. 

The main contributions of our work are listed as follows:
\begin{itemize}
	\item We propose a framework that integrates an active learning scheme with a deep learning model for video-based person re-ID. It performs re-ID model update and active annotation in an iterative and progressive way. In contrast to other semi-supervised or AL-based methods, our model requires no labeled re-ID data for initialization.
	\item We design a sampling criterion to choose the most TP-likely candidates for annotation. A view-aware strategy and an adaptive resampling step are also designed to facilitate candidate selection. Our sampling strategies can significantly reduce annotation effort.
	\item Extensive experiments on three benchmark multi-camera person re-ID datasets validate the effectiveness of the proposed method. The results show that less than 3\% pairwise annotations are needed for our method to reach comparable performance with the fully-supervised setting.  
\end{itemize}

\section{Related Work}
\subsection{Fully Supervised Video-based Person Re-ID}
The majority of existing video-based person re-ID methods are fully supervised. Similar to image-based counterparts, metric learning and representation learning are two major research directions. For instance, You \etal~\cite{You2016video} and Zhu \etal~\cite{Zhu2016metric} introduced set-based constraints into distance metric learning to better tackle intra-person variations in videos. McLaughlin \etal~\cite{McLaughlin2016video}, Zhou \etal~\cite{Zhou2017rnn}, and Li \etal~\cite{li2017video} designed recurrent neural networks (RNN) or pooling schemes to aggregate temporal features. Besides, attention schemes~\cite{Xu2017attent, chen2018video} were also introduced to the person re-ID problem in very recent years. Fully supervised methods have gained promising performances in large-scale video datasets~\cite{zheng2016mars, ristani2016performance}. However, their performances may degenerate dramatically when applied to real-world scenarios beyond the labeled training data domains.

\subsection{Semi-supervised Video-based Person Re-ID}
Semi-supervised learning trains a model initially on a small amount of labeled data and then update the model by exploiting unlabeled data. By this means, it can alleviate annotation burden without compromising too much performance. In semi-supervised video-based person re-ID, the one-shot setting~\cite{liu2017stepwise}, in which one tracklet of each identity is labeled, was considered in very recent years. For instance, Wu \etal~\cite{wu2018exploit} initialized a CNN model under the one-shot setting and gradually chose the most confident unlabeled tracklets for model update. DGM~\cite{ye2017dynamic} and the stepwise method~\cite{liu2017stepwise} also require at least one labeled tracklet for each identity to initialize their models. Different from active learning, these methods do not actively choose unlabeled data for human to annotate.

\subsection{Active Learning for Person Re-ID}
Active learning aims to reduce annotation cost by intelligently choosing some of the unlabeled data to annotate, and thus related to human-in-the-loop approaches~\cite{wang2016inloop}. Most existing works~\cite{Das2015active, Wang2016active, roy2018exploiting} applying active learning for person re-ID are based on still images. Different sampling strategies, such as the entropy-based criterion~\cite{Das2015active} and the exploration-exploitation jointed criterion~\cite{Wang2016active} that measures both diversity and uncertainty, were proposed. In~\cite{roy2018exploiting}, image-pair selection is formulated as a combinatorial optimization problem based on transitivity. All these image-based methods either work under the one-shot setting or require a small pre-labeled training set for model initialization. Contrastively, our work is video-based and no labeled person re-ID data is needed. Moreover, our proposed sampling strategy is quite different from theirs.

\section{The Proposed Method}
\subsection{Overall Framework}
Unlike many other active learning approaches~\cite{roy2018exploiting, Lin2018active}, the proposed method does not require any labeled data for initialization. Thus, we consider a fully unlabeled video dataset. By pedestrian detection and tracking, we get $C$ tracklets containing $N$ pedestrian images. Let us represent the dataset by $\mathcal{U} = \{x_1, \cdots, x_N\}$, where $x_i$ denotes the $i$-th image. $\mathbf{M}\in R^{N\times C}$ is a matrix mapping the image index to the tracklet index. If the $i$-th image belongs to the $j$-th tracklet, then the entry $\mathbf{M}_{ij} = 1$ and 0 otherwise.

Following~\cite{Zhong2017re-ranking, wu2018exploit}, we formulate the re-ID task as a classification problem that minimizes the following objective function:
\begin{equation}
\min_{w,\theta}\sum_{i=1}^{N} l(f(\mathbf{w}; \phi(\mathbf{\theta}; x_i)), y_i),
\label{equation-1}
\end{equation}
where $\phi$ is a CNN model, parameterized by $\mathbf{\theta}$, to extract the feature for the image $x_i$. $f$ is a classifier, parameterized by $\mathbf{w}$, to predict $N_c$-dimensional classification confidence. $N_c$ is the number of classes which is dynamically set. $l$ is the classification loss that computes the cross entropy between the prediction $f(\mathbf{w}; \phi(\mathbf{\theta}; x_i))$ and the pseudo target label $y_i$ that is automatically assigned in a certain way. 

\begin{figure*}[t]
	\centering
	\includegraphics[width=0.9\textwidth]{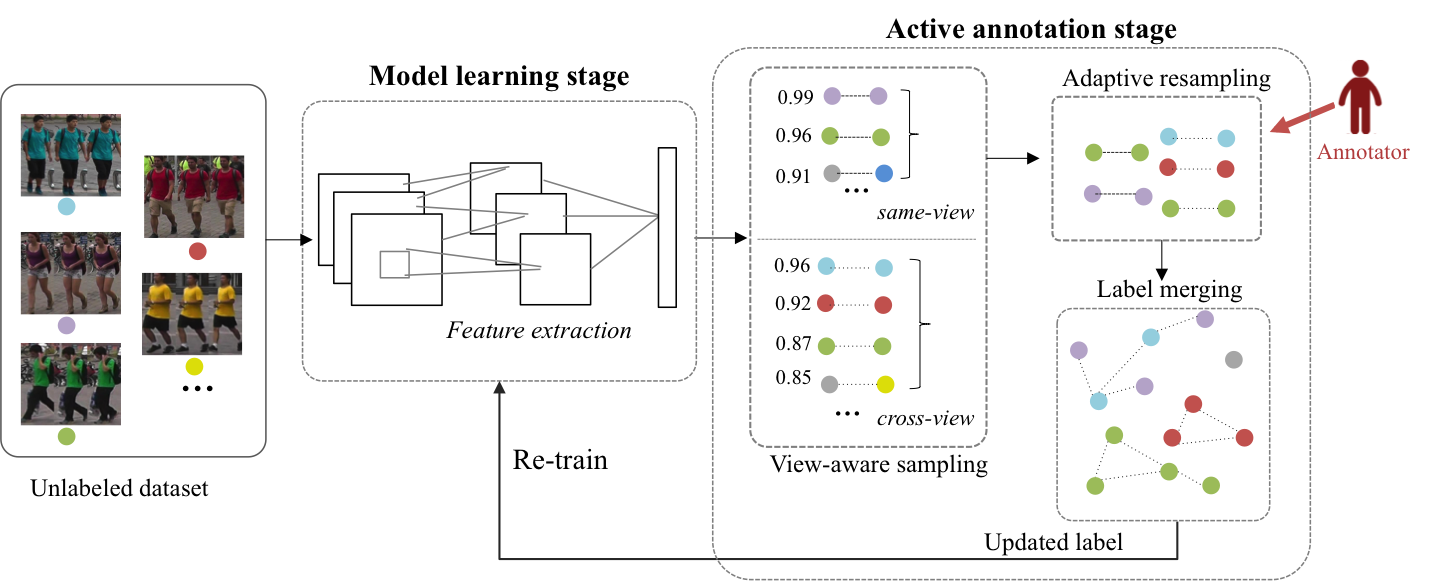}
	\caption{An overview of our proposed framework. The circles with different colors represent tracklets of different persons. Our model learns through iterating the two stages: model learning stage and active annotation stage. The pseudo image label is first initialized by considering each individual tracklet as belonging to a unique class. Using the updated pseudo image label as the target, the model learning stage learns under the supervision of the classification loss. Afterwards, the learned features are utilized by the active annotation stage for computing tracklet similarity. In the active annotation stage, the view-aware sampling strategy progressively selects the "TP-likely" tracklet pairs as candidates, then a re-sampling step performs label propagation so as to filter false positives. The chosen pairs are then annotated and merged into the updated target label for iterative model learning. The figure is best viewed in color.}
	\label{fig:pipeline}
\end{figure*}

Note that, for training efficiency, the above classification model takes each image as the input. In test stage, we use $\phi$ to extract the feature for each image of a query tracklet and gallery tracklets. A set-to-set distance defined in Sec.~\ref{sec:selection} is then applied to compute the distance between the query and gallery tracklets for the result ranking.  

The above-defined problem is optimized in an alternative way. In the beginning, each tracklet is treated as a distinct class. That is, the tracklets' pseudo target labels $\mathbf{Z} \in R^{C\times 1}$ are initialized as $\mathbf{Z}=[1, \cdots, C]^T$. According to the image-tracklet relation denoted in $\mathbf{M}$, the image's pseudo target label $y_i$ can be transitively obtained. Once initialized, we optimize $\mathbf{\theta}$ and $\mathbf{w}$ by fixing $y_i$, and then update $y_i$ by fixing the other parameters.

Fig.~\ref{fig:pipeline} presents an overview of our framework. Corresponding to the above-introduced optimization way, we split the entire procedure into model learning stage and active annotation stage that are performed alternatively. Our re-ID model adopts ResNet-50 \cite{he2016resnet} as the feature extractor, followed by several fully connected layers as the classifier. The feature extractor is pretrained not on any labeled re-ID data but only on ImageNet~\cite{imagenet_cvpr09}. The model learning stage jointly trains the feature extractor and the classifier under the supervision of the pseudo target labels. The active annotation stage first extracts image features using the learned feature extractor $\phi$, and then updates the pseudo labels by annotating tracklet-pairs actively selected according to a view-aware sampling and adaptive resampling strategy. After gaining incremental annotations, the tracklets'pesudo labels are furthered updated by a merging algorithm. At each new iteration, the number of classes $N_c$ in the re-ID model is reset as the number of merged clusters. 

\subsection{View-aware Sampling Strategy}
\label{sec:selection}
At each iteration, the active annotation stage selects the most informative tracklet-pairs for annotation. Manual annotation tells whether a selected pair is a true match or not. We observe that the true matches only take a small portion of whole pairs in a dataset, and the entire relationship between tracklets can be known if all true matches are annotated. Therefore, we prefer to choose the tracklet pairs that are the most likely to be true matches. To this end, we define a set-to-set distance as the criterion and design a view-aware strategy for sampling. Our view-aware sampling strategy is designed based on the view-specific biases introduced in Sec.~\ref{sec:introduction}. 


%

\textbf{The Set-to-Set distance.} 
The dissimilarity between tracklets is defined based on a set-to-set distance introduced here. Let us consider two tracklets $\mathcal{P}$ and $\mathcal{Q}$, each of which contains a set of pedestrian images. The distance between $\mathcal{P}$ and $\mathcal{Q}$ is defined by 
\begin{equation}
d(\mathcal{P}, \mathcal{Q}) = \min_{\mathcal{S}} \frac{1}{K}\sum_{ij} s_{ij}||\phi(\mathbf{\theta}; p_i) - \phi(\mathbf{\theta}; q_j)||_2,  
\label{eq:distance}
\end{equation}
Here, $p_i$ ($i\in\{1, \cdots, |\mathcal{P}|\}$) is an image belonging to tracklet $\mathcal{P}$, and image $q_j$ ($j\in\{1, \cdots, |\mathcal{Q}|\}$) is from $\mathcal{Q}$; $|\cdot|$ denotes the cardinality of a set. $s_{ij} \in \{0, 1\}$ is an indicator determining whether the distance between two images is counted in or not; $\sum_{ij} s_{ij} = K$, and $\mathcal{S} = {s_{ij}}$.

This distance takes the average of $K$ smallest image-pair distances as the distance between tracklets. In contrast to performing temporal pooling to obtain tracklet features and then compute the Euclidean distance between the tracklet features~\cite{wu2018exploit, ye2017dynamic, zheng2016mars}, or taking the average distance of all image-pairs~\cite{liu2017stepwise}, our approach is more robust to outliers. In our experiments, $K$ is set to be 3.

\textbf{View-aware sampling strategy.} The view-specific biases inspire us to design a sampling strategy that is aware of camera views. Specifically, at the $t$-th iteration, the active annotation stage selects $m_1(t)$ number of candidates that have the smallest dissimilarity values from the same-view tracklet pairs, together with $m_2(t)$ number of candidates from cross-views. For simplicity and efficiency, we set these two variables as follows: 
\begin{equation}
m_1(t)=\left\{
	\begin{array}{lr}
	s_1 &    {if\quad t<t_0} \\
	s_2 &    {otherwise},
	\end{array} \right.
\label{equation-m1}
\end{equation}
\begin{equation}
m_2(t)=\left\{
	\begin{array}{lr}
	s_3 &    {if\quad t<t_0} \\
	s_4 &    {otherwise}.
	\end{array} \right.
\label{equation-m2}
\end{equation}
The above setting indicates that we piecewise linearly increase the number of annotations along with the iterations going on. Considering that tracklet pairs from the same views are more similar to each other, we set a larger value for $s_1$ than $s_3$ so that more same-view pairs are selected for annotation at initial iterations. Later on, more cross-view pairs are selected by setting $s_4$ greater than $s_2$. This sampling strategy follows the self-paced learning principle by not only sampling in a progressive manner~\cite{wu2018exploit, wang2016human} but also shifting from same-view (easier) to cross-view (harder). Such self-paced manner can bring more reliability for a learner that has a weak initialization. 

\subsection{Adaptive Resampling}
The proposed sampling strategy is prone to choose tracklet-pairs that are more likely to be true matches. As iteration goes on, a growing number of the true-matches are being annotated, leaving only a small amount unfound. As a result, the percentage of selected true matches will decrease at later iterations and a lot of annotation efforts will be wasted on the selected false positive pairs. In order to reduce unnecessary annotations, we further propose an adaptive resampling scheme to leave out negative pairs selected at each iteration. 

Our adaptive resampling scheme is designed by first using an efficient label propagation technique~\cite{zhu2002lableprop} to propagate clumped clusters to isolated ones, and then using a reciprocal ranking rule to filter out negatives. We briefly introduce these two step as follows.

We consider all the tracklets in a dataset and the pesudo target labels $\mathbf{Z}$ that is defined previously. The labels are soft labels that can be interpreted as distributions over clusters. We let the labels of a tracklet propagate to all other tracklets through fully connected edges. A probabilistic transition matrix $\mathbf{T} \in R^{C\times C}$ is defined by~\cite{zhu2002lableprop}:
\begin{equation}
\mathbf{T}_{ij} = \frac{w_{ij}}{\sum_{k}w_{kj}},
\label{LP}
\end{equation}
where 
\begin{equation}
w_{ij} = \exp(\frac{-d_{ij}^2}{\sigma}),
\end{equation}
is a weight computed according to the tracklet distance $d_{ij}$ between the $i$-th tracklet and the $j$-th tracklet.

The label propagation technique~\cite{zhu2002lableprop} propagates the distributions between all tracklets by iteratively perform the following steps:
\begin{enumerate}
	\item $\mathbf{Z} \longleftarrow \mathbf{T}\mathbf{Z}$;
	\item Row-normalize $\mathbf{Z}$;
	\item Clamp the results.
\end{enumerate}
 
After label propagation, we can derive the probability distribution of each tracklet belonging to each cluster, which is further used to screen candidates. We adopt the reciprocal ranking as a rule for screening. Assume $\mathcal{S}$ is the set of candidate tracklet-pairs obtained from the sampling stage. Assume $\mathcal{N}_K(i)$ is the $K$-nearest cluster neighbors of tracklet $i$, i.e. the top-K of the ranked probability distribution after performing label propagation. Then 
\begin{equation}
\mathcal{S}_f=\{(t_1,t_2)|t_1  \in \mathcal{N}_K(t_2)\wedge t_2 \in \mathcal{N}_K(t_1); (t_1, t_2) \in \mathcal{S}\}
\label{re-selection}
\end{equation}
denotes the candidate pairs remained after screening. The rule in~\ref{re-selection} indicates that a tracklet-pair is kept when both tracklets in the pair are among the $K-$nearest cluster neighbors of each other. Otherwise the pair is removed from the candidate set.


\subsection{Label Merging}
Our re-ID model is iteratively trained with the supervision of all tracklets' pseudo target labels. These pseudo labels are initialized by taking each tracklet as a distinct class. After receiving annotations for the progressively sampled tracklet-pairs, we take a label merging process at each iteration to reduce the class number for the re-ID model. The merging result is required to satisfy 1) each tracklet in a cluster should be matched with one or more other tracklets in the same cluster and 2) a tracklet outside a cluster is not matched with any tracklets in the cluster. 

We adopt a density-based clustering algorithm DBSCAN~\cite{ester1996density} for merging. DBSCAN basically groups together the points in high density and marks the points that lie alone in low-density areas as outliers. Therefore, it can discover clusters of arbitrary shape in spatial databases with noise. There are two key parameters in DBSCAN: $\epsilon$ and $\mathcal N(P)$ that, respectively, denotes the radius for the neighborhood of a point $P$ and the minimum number of points in the given neighborhood $\mathcal N(P)$. In our implementation, we set $\epsilon$ to 0.01 and $MinPts$ to 2, so that our requirements can be satisfied to a large extent.

The merged labels provide a pair-consistent~\cite{paul2017non, zhou2017fine, Lin2018active} picture for all tracklet-pairs. For instance, a tracklet in a cluster matches to all the other tracklets in the same cluster. Moreover, if two tracklets from separate clusters are identified, then these two clusters can be merged into one cluster. These consistencies bring a lot of auto-annotated pairs and boosts the annotation efficiency.

\section{Experiments}
\subsection{Datasets}
\textbf{The PRID dataset}~\cite{hirzer2011person} consists of images captured by two cameras, with 385 identities recorded by one camera, and 749 identities by the other. 200 identities appear in both camera views. In order to guarantee the effective length of videos, 178 identities each of which has more than 27 frames are selected out of the mutual 200 identities. During experiments, the dataset is randomly divided by half into training and test sets. The train/test partition are repeated 10 times and the average results are reported.

\textbf{The MARS dataset}~\cite{zheng2016mars} is the largest video dataset for person re-ID. It contains 20,478 tracklets for 1,261 identities, captured by six cameras on a university campus. The tracklets are automatically generated by the DPM~\cite{felzenszwalb2010object} detector and the GMMCP~\cite{ristani2016performance} tracker. The dataset is evenly split into training and test sets, respectively, containing 631 and 630 identities. We fix this partition in our experiments. During test, each identity has one randomly-selected tracklet probe under each camera. 
 
\textbf{The DukeMTMC-VideoReID (Duke-video) dataset} is a recent video re-id dataset, created by Yu \etal~\cite{wu2018exploit} in their experiments for one-shot person re-id. It is a subset of DukeMTMC dataset~\cite{ristani2016performance}, a large-scale dataset for multi-camera tracking. The tracklets are generated by cropping pedestrain images from the videos for 12 frames every second. Since the DukeMTMC dataset is manually annotated, each identity has at most one tracklet under each camera. Following the protocol in~\cite{zheng2017unlabeled}, the generated DukeMTMC-VideoReID dataset is split into 702 identities for training, 702 identities for test and 408 identities as distractors. 

\subsection{Experimental Settings}
\textbf{Evaluation metrics.} For both MARS and DukeMTMC-VideoReID, the Rank-1 score of the cumulative matching characteristic (CMC) curve and the mean average precision (mAP) are adopted to measure the re-id performance. For the PRID dataset, since each query has only one ground truth, we report the Rank-1, Rank-5, Rank-10, Rank-20 scores of the CMC curve.

\textbf{Annotation ratio.} The annotation ratio (AR) is defined as the number of labeled data divided by the number of whole data. Due to different annotation ways existing in re-ID works, we here provide the exact definition. Let us denote the number of identities in a dataset by $I$ and the number of tracklets by $C$. For the methods that annotate tracklet-pairs like ours, the annotation ratio can be computed as $AR=\frac{tp}{T_{pa}}$, where $tp$ is the number of manually annotated tracklet-pairs, and $T_{pa}$ is the total number of pairs needed to label the whole dataset. $T_{pa}$ can be computed as follows: Under common annotation settings, pairs are randomly selected for labeling. For a newly annotated pair, the historical annotation information of the involved tracklets are synchronized between them. Since there is no direct formula to compute the above described $T_{pa}$, we perform intensive simulations and use the average total annotation result as $T_{pa}$.
For the methods that directly assign an ID to each tracklet, if $t$ tracklets are labeled, then $AR = \frac{t}{C}$.

\textbf{Implementation details.} The proposed method is implemented using the PyTorch~\cite{paszke2017automatic} framework. During training, the batch size is set to 32 for MARS and DukeMTMC-VideoReID dataset, and 8 for PRID since the last dataset is relatively small. We use stochastic gradient descent (SGD) as the optimizer with weight decay 0.9 and momentum 5e-4. The learning rate is fixed to 0.001 in our experiments. 

\subsection{Algorithm analysis}

\begin{figure*}[t]
\begin{minipage}[t]{0.32\textwidth}
\centering
\includegraphics[width=1\textwidth]{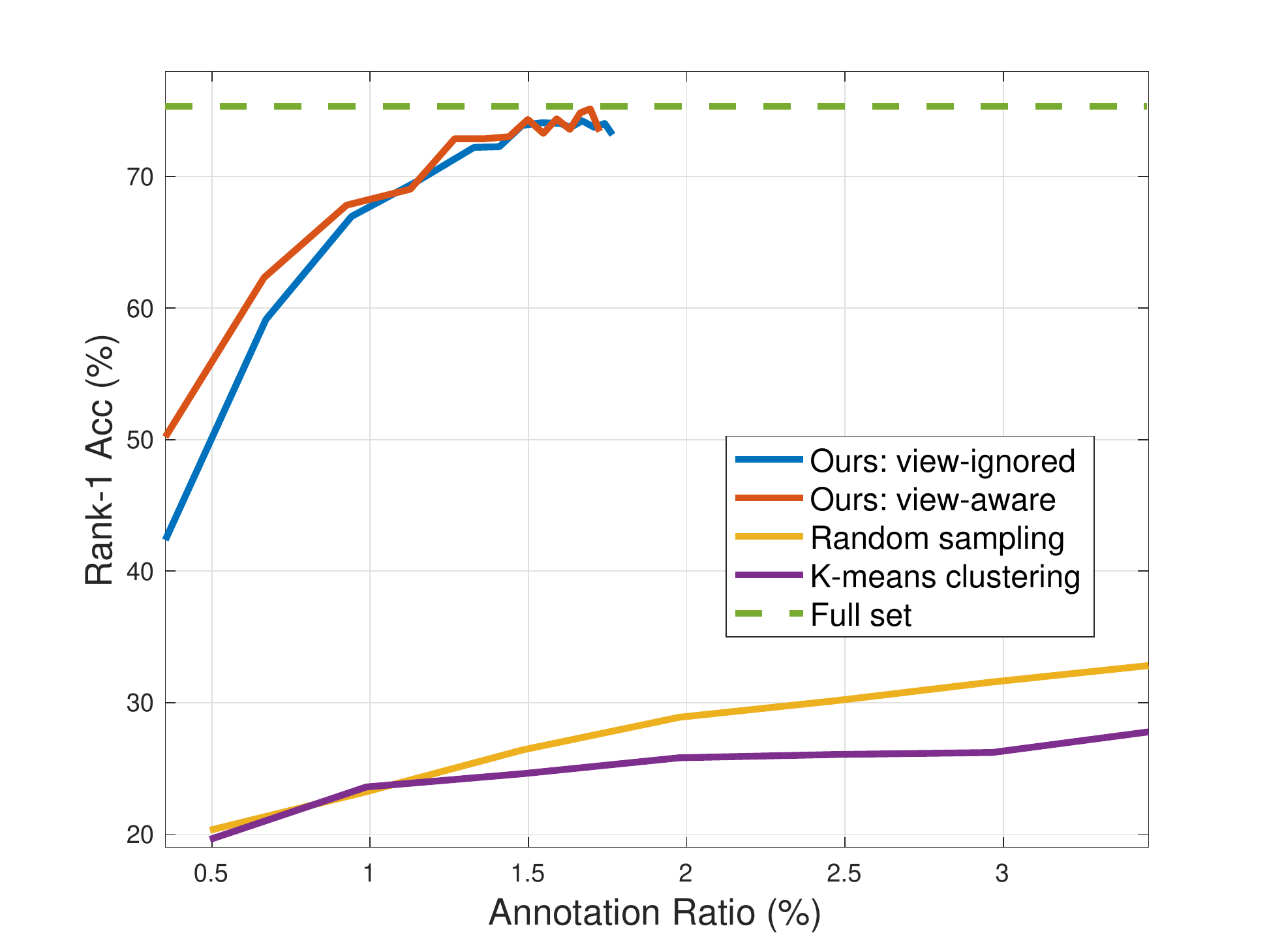}
\caption*{(a)}
\end{minipage}
\quad  
\begin{minipage}[t]{0.32\textwidth}
\centering
\includegraphics[width=1\textwidth]{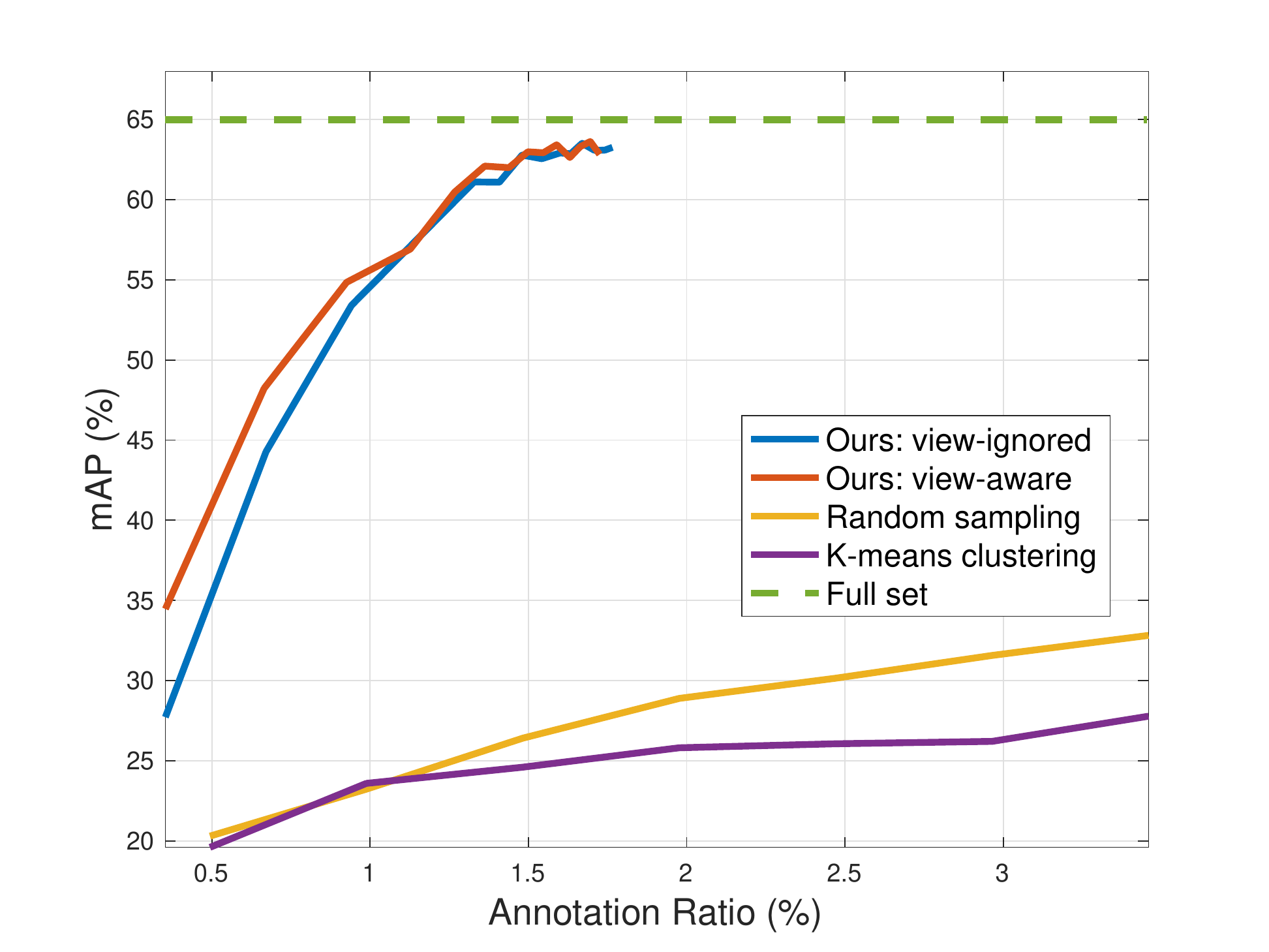}
\caption*{(b)}
\end{minipage}
\quad  
\begin{minipage}[t]{0.32\textwidth}
\centering
\includegraphics[width=1\textwidth]{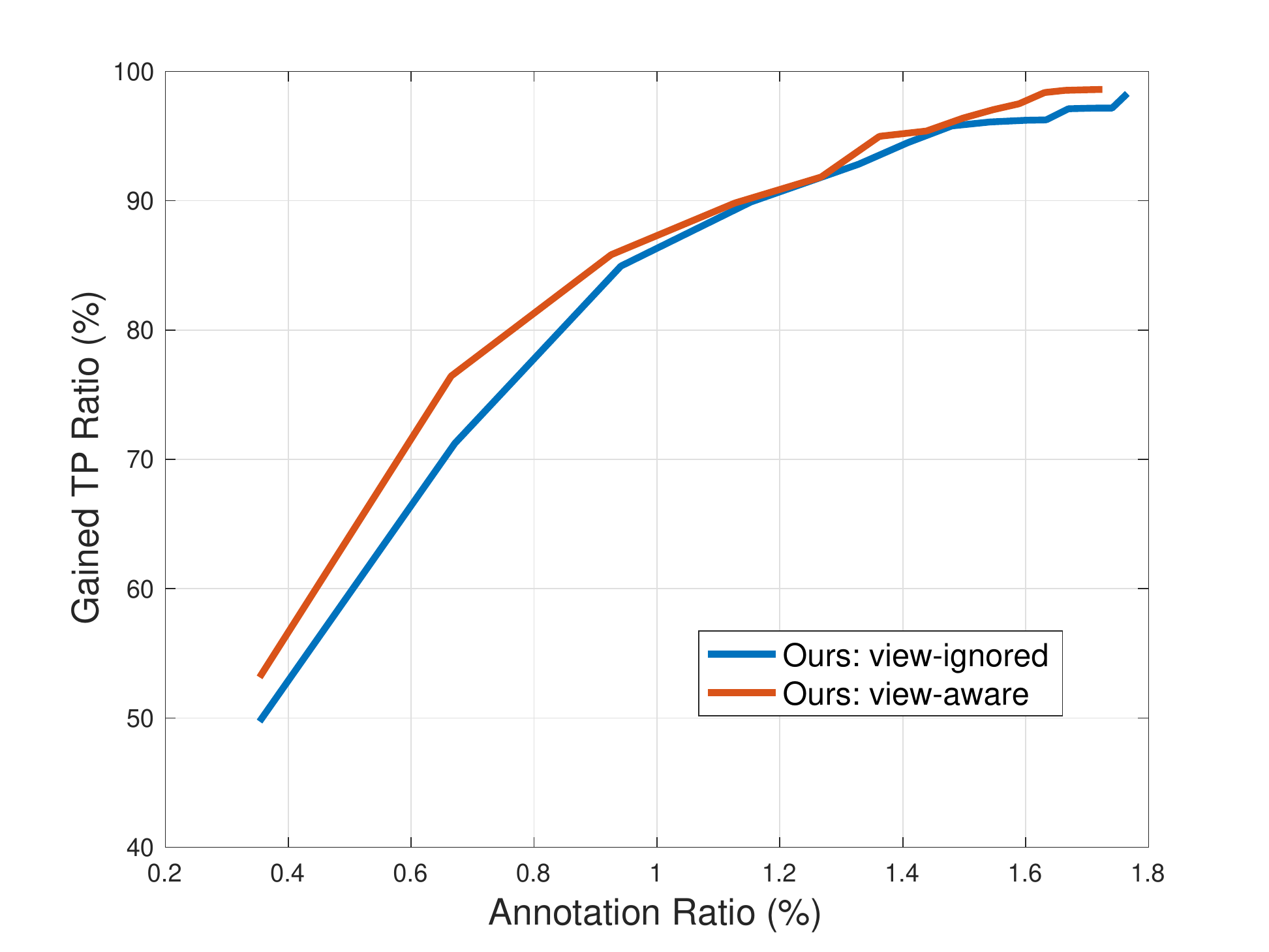}
\caption*{(c)}
\end{minipage}
\\
\begin{minipage}[t]{0.32\textwidth}
\centering
\includegraphics[width=1\textwidth]{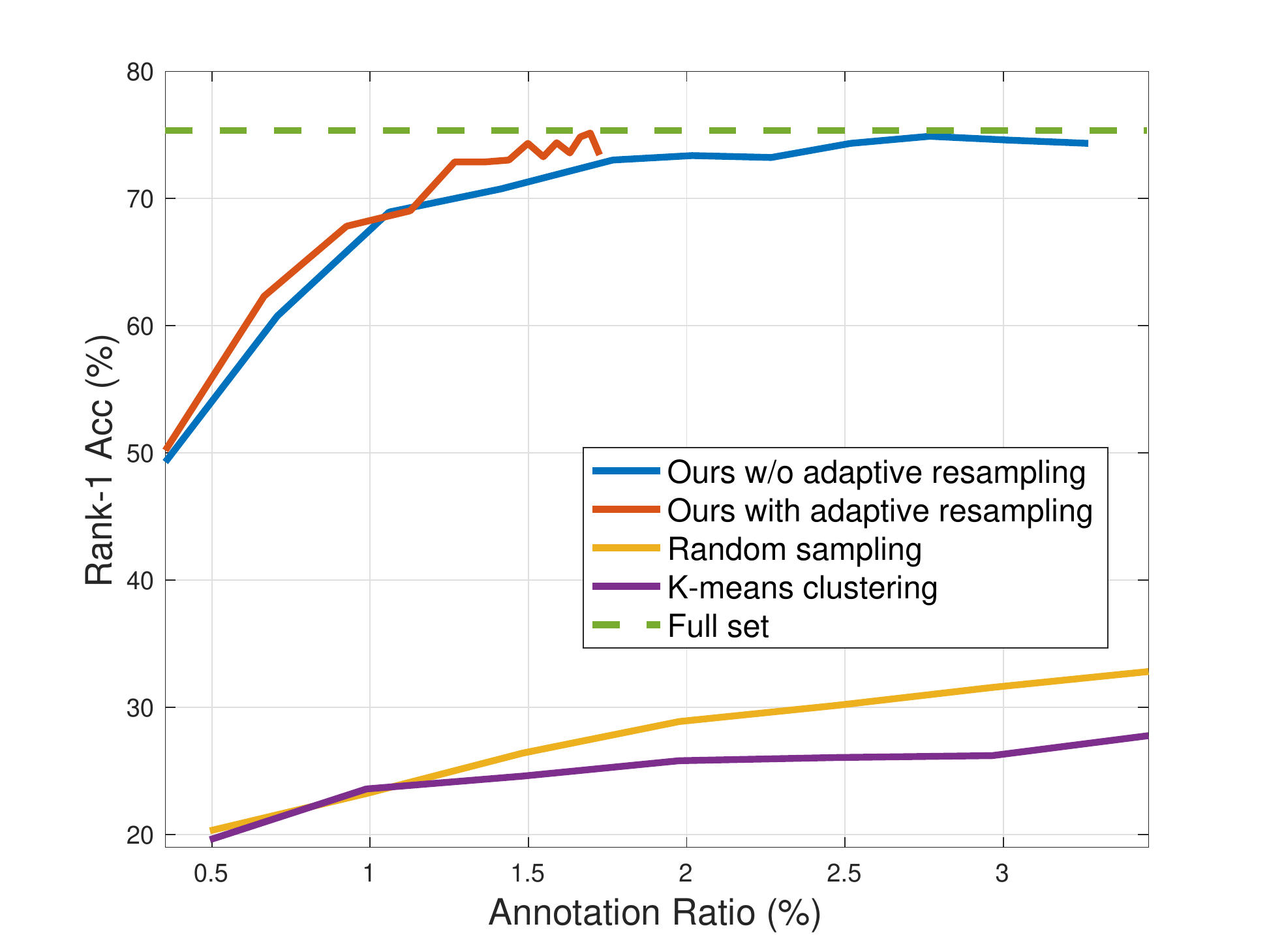}
\caption*{(d)}
\end{minipage}
\quad  
\begin{minipage}[t]{0.32\textwidth}
\centering
\includegraphics[width=1\textwidth]{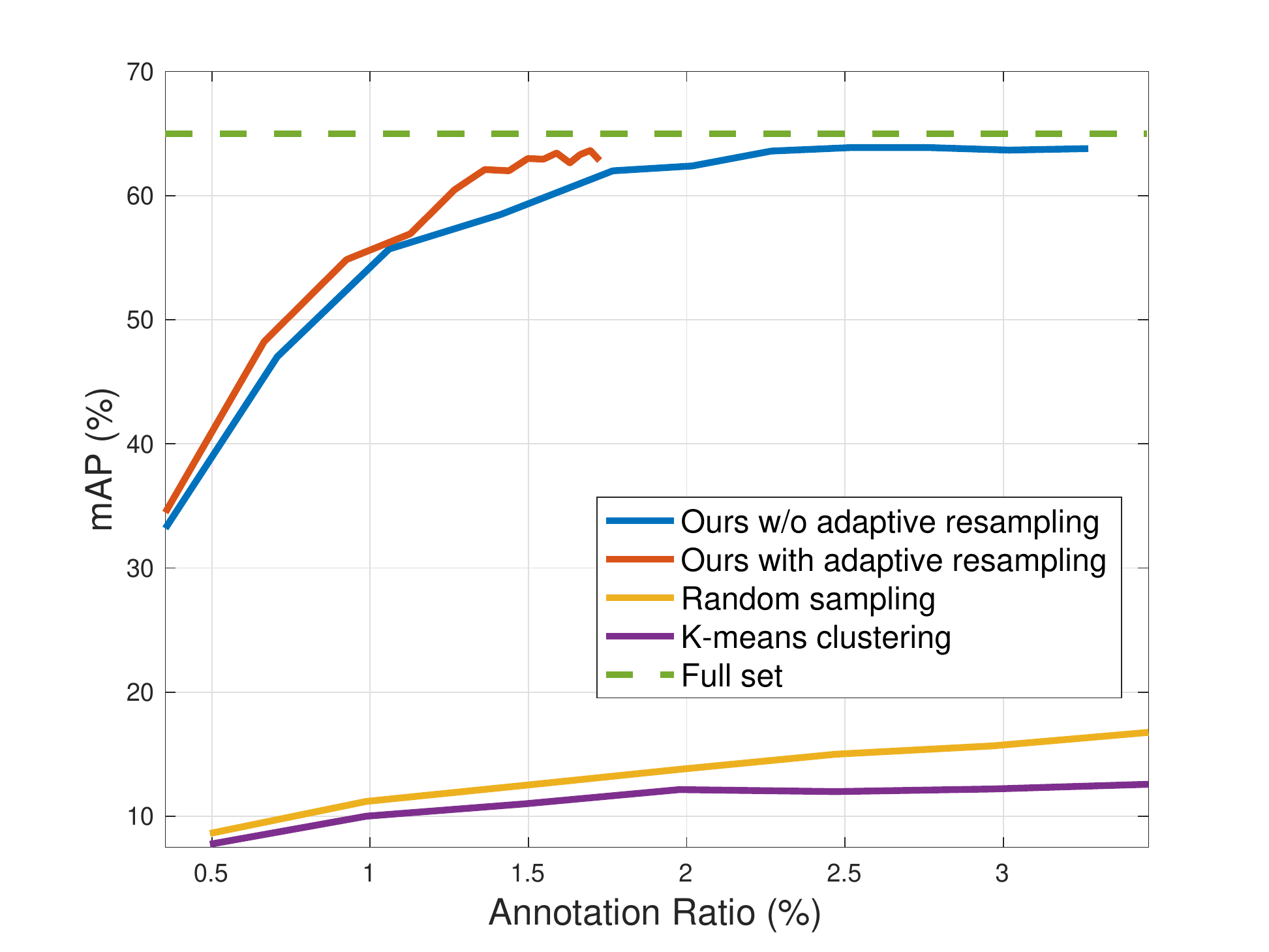}
\caption*{(e)}
\end{minipage}
\quad  
\begin{minipage}[t]{0.32\textwidth}
\centering
\includegraphics[width=1\textwidth]{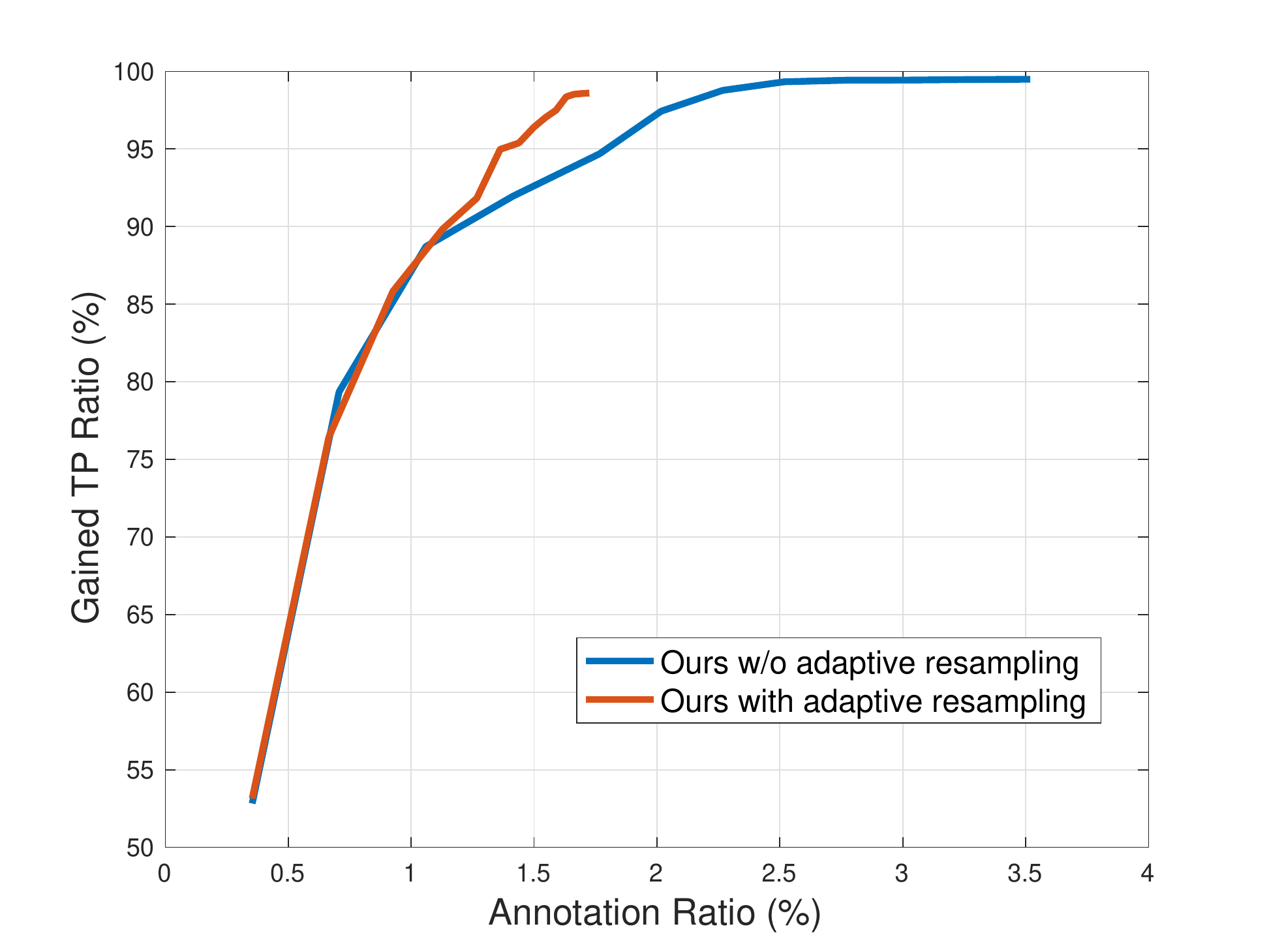}
\caption*{(f)}
\end{minipage}
\caption{Ablation comparison on the view-aware sampling strategy and the adaptive resampling strategy. (a),(b) and (c) are for comparing the view-aware sampling strategy: (a) and (b) show the rank-1 accuracy and the mAP curve over the annotation ratio, respectively. (c) plots the gained TP ratio as the annotation ratio increases. (d),(e) and (f) are for comparing the adaptive resampling strategy: (d) and (e) show the rank-1 accuracy and the mAP curve over the annotation ratio, respectively. (f) plots the gained TP ratio as the annotation ratio increases.
}	
\label{ablation_comparisons}
\end{figure*}
The proposed framework consists of several key components that altogether contribute to the final performance. In order to investigate how these individual components influence the model performance, we conduct the following ablation experiments on MARS dataset. 

\subsubsection{Analysis on view-aware sampling strategy.}
The view-aware sampling strategy splits all the pairs into two subsets, according to whether the tracklets come from the same camera view or not. Then candidate pairs are progressively selected from each of the two subsets. The main advantage of such design is that the difference of pair hardness can be considered. To investigate how this strategy contribute to the performance, we compare 1) the mode with the view-aware strategy2) the model that treats all views equally, as well as other active learning methods for re-ID. 

The tradeoff between re-ID accuracy and manual annotation ratio is illustrated in Fig.~\ref{ablation_comparisons}. In specific, Fig.~\ref{ablation_comparisons}(a) and Fig.~\ref{ablation_comparisons}(b) show the rank-1 accuracy and mAP over the manual annotation ratio, respectively. Fig.~\ref{ablation_comparisons}(c) plots the relationship between the manual annotation ratio and the actual gained TP ratio. From Fig.~\ref{ablation_comparisons}(a), we can see that both the two candidate selection strategies are able to reach the fully-annotated rank-1 accuracy with a tiny amount (less than 1.2\%) of annotations, which greatly outperforms the random sampling and the k-means clustering methods. In addition, the resampling strategy consistently outperforms the mixed strategy. For example, when the annotation ratio is 0.23\% (i.e. around 12000 tracklet pairs), the resampling strategy achieves 50.2\% rank-1 accuracy, surpassing the mixed-strategy by 7.9\% (absolute). The mAP curves in Fig.~\ref{ablation_comparisons}(b) are in accordance with the rank-1 curves in Fig.~\ref{ablation_comparisons}(a). When the annotation ratio is greater than 0.75\%, the superiority of resampling strategy over mixed-view declines, since at this time most of the TP pairs are annotated, causing the difference to be less significant. Nevertheless, the advantage of view-aware over view-ignored can be well proved in general. 

The curves shown in Fig.~\ref{ablation_comparisons}(c) can further explain the accuracy curve behaviors. The gained TP ratio means the percentage of the gained TP pair number to the total TP number. It can be observed in Fig.~\ref{ablation_comparisons}(c) that as the annotation ratio increases, the gained TP ratio first rises rapidly, and then slowly reaches near-100\%. It indicates that the gained TP number is the key factor to the improvement of recognition accuracy.

\subsubsection{Analysis on adaptive resampling.}
To make better use of the annotation resource, we propose an adaptive resampling scheme to further filter out the selected negative candidates. In order to analyze the effect of this scheme, we conduct the following experiments with and w/o resampling for explicit comparison. The experimental results are presented in Fig.~\ref{ablation_comparisons}. 

The rank1 and mAP curves vs annotation ratio are shown in Fig.~\ref{ablation_comparisons}(d) and (e), and performances of the baseline active learning methods are compared as well. Several conclusions can be inferred from the two sub-figures: First, we can observe the consistency between rank-1 and mAP curve trends, as well as the large performance gap between our proposed model and the two compared active learning methods. In addition, our model with adaptive resampling almost always performs better than the model w/o it, reaching higher rank-1 and mAP accuracy when under the same annotation ratio. Last but not least, the model with adaptive resampling reaches fully-supervised performance using much less annotations, which proves the effectiveness of the adaptive resampling step at removing false positive pairs. For better analysis, we also present the relationship between the gained TP ratio and the manual annotation ratio in Fig.~\ref{ablation_comparisons}(f). As is shown in Fig.~\ref{ablation_comparisons}(f), the two settings (with and w/o resampling) have quite similar TP gains in the beginning. As iteration goes on, the percentage of TP pairs in the candidates starts to fade while the the percentage of FP pairs is on the increase. At this time, the effect of the resampling gets more significant. Finally, the setting with resampling is able to discover almost all the TPs at a lower manual annotation ratio.

\begin{table*}[ht]
\label{AL_compare_all}
\centering
\scalebox{0.94}{
\begin{tabular}{c|l|c|c|c|c|c|c|c|c|c|c|c}
\hline  
\multirow{2}{*}{Type}
& \multirow{2}{*}{Method} & \multicolumn{5}{c|}{PRID} & \multicolumn{3}{c|}{MARS} 
& \multicolumn{3}{c}{Duke-video}\\
\cline{3-13}& & A.R. & R1 & R5 & R10 & R20 & A.R. & R1 & mAP & A.R. & R1 & mAP\\ \hline
\multirow{4}{*}{Supervised}
&caffeNet\cite{zheng2016mars} &100&77.3& 93.5&$-$&99.3&	100& 65.3&	47.6& $-$& $-$&	 $-$\\
&Fusion\cite{li2017learning}  &$-$& $-$& $-$& $-$& $-$& 100& 83.03& 66.43& $-$& $-$&	 $-$\\
&GOG+XQDA\cite{matsukawa2016hierarchical}  & 100& 69.4& 89.6&	92.4& 95.7&	100& 41.97&	24.89&	100& 58.83&	52.42\\
&Ours(supervised)  &100 &73.93&	88.31&	92.36& 96.63&  100& 75.35& 64.98&  100& 87.04& 83.46\\ \hline
\multirow{4}{*}{One-shot}
&SMP\cite{liu2017stepwise} & 50& 38.7& 68.1& 79.6& 90.0&  7.53& 41.2& 19.7&	31.97& 56.26&	46.76\\
&DGM\cite{ye2017dynamic} & 50&  48.2&  78.3& 83.9& 92.4&  7.53& 36.8& 21.3&	31.97& 42.36&	33.62\\
&EUG\cite{wu2018exploit} & $-$& $-$&  $-$&  $-$&  $-$&   7.53& 62.67& 42.45& 31.97& 72.79&  63.23\\
&RACE\cite{ye2018robust} & 50&  50.6& 79.4&	84.8& 91.8&	 7.53& 43.2& 24.5&	$-$& $-$&   $-$\\ \hline
\multirow{3}{*}{Active learn}
&Random sample & 50.56& 44.49& 71.01& 80.11& 90.56&  1.98&	28.89&	13.78&	34.15&  54.56&	49.81\\
&K-means\cite{nie2013early} & 50.56& 52.7& 75.73&  82.81& 89.89&  1.98& 25.81&  12.15&  34.15&  61.54&  55.85\\
&\textbf{Ours}  &  \textbf{2.05}& \textbf{71.68}&  \textbf{89.44}&	\textbf{93.03}&	\textbf{95.84}&	\textbf{1.62}& \textbf{75.15}&	\textbf{63.62}&	\textbf{0.26}& \textbf{85.19}&	\textbf{80.11}\\ \hline
\end{tabular}
}
\caption{Performance comparison with other methods on PRID, MARS and Duke-video dataset. A.R. means the manual annotation ratio in percentage.}
\label{AL_compare_others}
\end{table*}

\subsection{Comparison with the State-of-the-Art Methods}
To validate the effectiveness of the proposed approach, we compare it to other deep learning based methods on all three datasets. These compared methods are grouped into three categories: 1) Fully supervised methods, including caffeNet~\cite{zheng2016mars}, Fusion~\cite{li2017learning}, Snippet~\cite{chen2018video}, and the supervised version of our approach. 2) Semi-supervised methods, including EUG~\cite{wu2018exploit}, DGM~\cite{ye2017dynamic}, SMP~\cite{liu2017stepwise}, and RACE~\cite{ye2018robust}. These methods are learned under the one-shot setting. 3) Baseline active learning methods, including a random sampling method and a K-means clustering approach~\cite{nie2013early}. The former is the version using our framework but replacing the sampling strategy by random sampling. K-means clustering ranks the samples by their distances to the K cluster centers in ascending order, and selects the top-ranked samples for ID annotation.

Table~\ref{AL_compare_others} reports the comparison results on three datasets. From Table~\ref{AL_compare_others}, we can make the following observations:
\begin{itemize}
	\item With less than 3\% annotations, our approach reaches comparable performance to our fully-supervised counterpart on all three datasets. Specifically on MARS, our approach achieves 98.67\% of the fully-annotated counterpart with only 1.13\% annotations. In comparison, the active learning methods using random sampling or k-means clustering give significantly worse results while using more annotations. The comparisons demonstrate the effectiveness of our method at wisely querying samples to reduce annotation amount.
	\item Since the one-shot methods require at least one annotated tracklet of each identity, their annotation ratio is computed as the percentage of labeled tracklets among all the tracklets. Hence the annotation ratio is 7.53\%, 31.97\% and 50\% for MARS, Duke-video and PRID respectively. When comparing with the one-shot methods, our method outperforms a lot with much less annotations required. The one-shot methods exploit the annotations better than random annotation, however their performance boost may be limited by the one-tracklet per-ID annotation requirement. On the other hand, the results show that a well-designed annotation strategy can better make use of the annotation amount to help promote re-id performance.
	\item When comparing with the fully supervised methods, our fully-supervised counterpart gives better results than GOG+XQDA \cite{matsukawa2016hierarchical}, and performs on par with caffeNet\cite{zheng2016mars}. The comparisons prove that our fully-annotated counterpart is an effective upper bound to verify our proposed active learning re-id method.
\end{itemize}

\section{Conclusion}
Reducing annotation cost is an important goal pursued by various computer vision applications. In this paper, we have presented a video-based person re-ID framework that integrates an active learning strategy to progressively select the most TP-likely tracklet-pairs for annotation. In our incremental selection process, a view-aware sampling strategy is adopted that takes view-specific biases into account to facilitate candidate selection. To further tackle the increasing number of selected negative pairs that are not necessary to annotate, we proposed an adaptive resampling step which effectively filters them out. The proposed approach has been validated on three public datasets. It reaches comparable re-ID performance to the fully-supervised setting while using an extremely low annotation amount. The experimental results demonstrate the effectiveness of our method. Being simple and flexible, our active learning strategy can be combined with other state-of-the-art deep re-ID networks to bring further improvement in re-ID performance and annotation efficiency.


{\small
\bibliographystyle{ieee}
\bibliography{reid_1_}
}

\end{document}